\pdfoutput=1

\documentclass[11pt]{article}

\usepackage{EMNLP2022}

\usepackage{times}
\usepackage{latexsym}
\usepackage{graphicx}
\usepackage{ragged2e}
\usepackage{float}
\usepackage{longtable}
\usepackage{graphics}

\usepackage[T1]{fontenc}

\usepackage[utf8]{inputenc}

\usepackage{microtype}

\usepackage{inconsolata}

\title{COMET-QE and Active Learning for Low-Resource Machine Translation}

\author{Everlyn Asiko Chimoto \\
  University of Cape Town, South Africa \\
  African Institute for Mathematical\\ Sciences \\
  \texttt{everlyn@aims.ac.za} \\\And
  Bruce A. Bassett \\
  University of Cape Town, South Africa \\
  African Institute for Mathematical\\ Sciences, South Africa \\
  South African Astronomical Observatory\\
  \texttt{bruce.a.bassett@gmail.com } \\}

\begin{document}
\maketitle
\begin{abstract}
Active learning aims to deliver maximum benefit when resources are scarce. We use COMET-QE, a reference-free evaluation metric, to select sentences for low-resource neural machine translation. Using Swahili, Kinyarwanda and Spanish for our experiments, we show that COMET-QE significantly outperforms two variants of Round Trip Translation Likelihood (RTTL) and random sentence selection by up to 5 BLEU points for 20k sentences selected by Active Learning on a 30k baseline. This suggests that COMET-QE is a powerful tool for sentence selection in the very low-resource limit. 
\end{abstract}

\section{Introduction}

Active Learning (AL) is a technique that involves querying an Oracle for labels based on a selection strategy. For Neural Machine Translation (NMT), active learning involves selecting sentences for translation based on estimated model uncertainty on the sentence or the diversity of the sentence. The aim of AL is to choose sentences that would maximally improve the NMT model. This is particularly attractive for low-resource languages where budgets are small, which includes most of Africa's languages which have historically been neglected \cite{nekoto-etal-2020-participatory}. 

There have been several advances in Active Learning for NMT with  Round Trip Translation Likelihood  (RTTL) being the current state of the art~\citep{zeng-etal-2019-empirical,peris-casacuberta-2018-active,zhao-etal-2020-active,DBLP:journals/corr/abs-2201-05700}. These advancements have typically been tested by artificially restricting high-resource languages which mimic true low-resource languages. There have been few  empirical evaluations of how these AL techniques perform on a low-resource African language with less than 100K sentences. We tackle this problem  while simultaneously proposing a new AL sentence selection strategy based on COMET-QE \citep{rei-etal-2020-unbabels} and show that it outperforms two versions of RTTL, one of the most successful AL selection algorithms, on Swahili, Spanish and Kinyarwanda. Interestingly the RTTL variants perform worse than random sentence sampling in this context. 

\section{Active Learning for NMT}

\begin{figure}
\centering
\resizebox{\columnwidth}{!}{
		\includegraphics[height=0.2\textheight,width=0.5\textwidth]{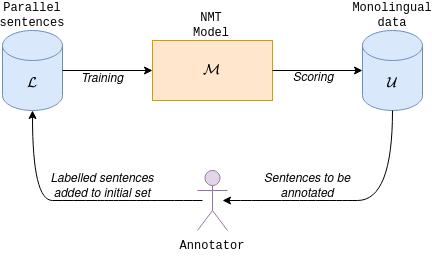}
}
\caption{Active Learning framework where the labelled set $\mathcal{L}$ is used to train model $\mathcal{M}$ which score monolingual sentences based on an uncertainty measure. After scoring, the sentences selected based on the annotation budget get translated and added to the training set for retraining.}
		\label{fig:AL}
\end{figure}

There are several AL sentence selection frameworks including pool-based AL, membership query synthesis and stream-based AL~\cite{Settles2009ActiveLL}. We focus on pool-based AL where we assume their exist a large pool of monolingual sentences for the language. Pool-based AL in NMT is an iterative process that involves selecting batches of sentences after each given AL cycle~\citep{haffari-etal-2009-active,bloodgood-callison-burch-2010-bucking}. Given a set of parallel sentences $\mathcal{L}$, a model $\mathcal{M}$ is trained. $\mathcal{M}$ is then used to translate monolingual data, $\mathcal{U}$ and a selection strategy scores and ranks the sentences in $\mathcal{U}$. The top sentences are then selected for translation. After translation, the selected sentences are added to set $\mathcal{L}$ and used to retrain and improve the model~\citep{ambati-etal-2010-active}; see Figure \ref{fig:AL}. This process is repeated until a specified stopping criteria is fulfilled, e.g. total translation budget reached.

The selection strategy aims to choose the N sentences that will improve the model the most. Selection strategies are typically either model-driven or data-driven. In Model-driven techniques a model scores sentences whereas Data-driven methods, also known as model-agnostic, involve comparing the set of labelled and unlabelled vernacular sentences to score the sentences. The current State-of-the-art for model-driven selection is RTTL \cite{zeng-etal-2019-empirical,DBLP:journals/corr/abs-2201-05700} which we discuss now. 

\subsection{Round Trip Translation Likelihood (RTTL)}
\citet{zeng-etal-2019-empirical} proposed RTTL as an AL selection strategy. RTTL involves training 2 models: i) Model $\mathcal{M}$ that translates from the source to target language and ii) Model $\mathcal{M}_{rev}$ that reverse translates back to the source language. Given a source sentence $x$, $\mathcal{M}$ generates a translation $\hat{y}$. $\mathcal{M}_{rev}$ translates $\hat{y}$ back to the source language. However, instead of  comparing of $x$ and $\hat{x}$, the probability of getting the actual sentence $x$ can be used to select sentences for translation. In RTTL, the key likelihood is defined in terms of $x$ and $\hat{y}$ as:
$$\phi(x,\mathcal{M},\mathcal{M}_{rev}) = -\frac{1}{L} \log P_{rev}(x|\hat{y})$$ 
where $L$ is the length of the sentence and $P_{rev}$ is the probability generated by the reverse model which estimates the probability of the original sentence given the translated sentence. The lower the probability the more uncertain the model is in translating the sentence and the more likely it will thus be selected for translation. We refer to this version as RTTL and our variant, discussed below.  

\subsection{Random Baseline} 

In addition to RTTL we compare our new sentence selection methods against a random baseline in which sentences are selected at random.

\section{New Active Learning Algorithms}
In this section, we detail our proposed new sentence selection techniques.

\subsection{COMET-QE }

Estimating the quality of a translation is a key area in evaluating machine translation. There have been several advancements in estimating the quality of a translation without having the reference or label. These reference-free metrics can be used as a selection strategy. Just as with RTTL, if the quality of a model translation is poor, this sentence is a good candidate for selection by active learning. 

COMET-QE is a reference-free metric proposed by \citet{rei-etal-2020-unbabels} under the WMT 2020 Shared Task on Metrics. COMET-QE correlates well to human assessments of translation quality. 

\begin{figure}
	\centering
    \resizebox{\columnwidth}{!}{
		\includegraphics[height=0.25\textheight,width=0.35\textwidth]{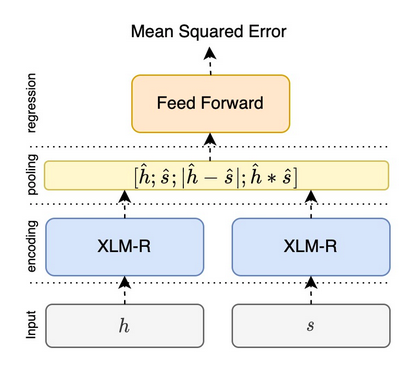}
		}
		\caption{The COMET-QE architecture. The model receives a source sentence, \textit{s} and a hypothesis, \textit{h}, passes them through the pre-trained XLM-Roberta model and performs regression to estimate a quality score. Figure source: ~\cite{Unbabel}.}
		\label{fig:comet}
\end{figure}

COMET-QE uses a pre-trained model, XLM-RoBERTa (XLM-R) ~\citep{conneau-etal-2020-unsupervised} to encode both the source sentences as well as the translation. Thereafter, a pooling layer converts the embeddings into segment-level vectors and finally the last layer produces a quality score; see Figure \ref{fig:comet}. 

For our current work, Swahili, Spanish and Kinyarwanda source sentences and English hypotheses were fed into the COMET-QE model, generating quality estimates for each  pair of  sentences. This AL algorithm then selects the lowest quality candidate sentence pairs for translation by the annotator. 

\subsection{Stratified RTTL}
Since RTTL selects sentences with low likelihood given the reverse translation model, one may be concerned that it selects overly long/short sentences that do not truly add diversity to the training dataset. 
Thus we propose stratified RTTL (S-RTTL) which implements RTTL on length-based stratification of the test or training sets. 

S-RTTL looks at the distribution of sentence length in the test or training set and selects the highest scoring sentences in each length bin/stratum proportional to the number of test/training sentences with that length. This ensures that RTTL is not biased to the sentences of a certain length. In effect, given the initial training data distribution, S-RTTL would select sentences that fall under the same length distribution of test/training. An example, if the training data has 10\% sentences of length 1-10, S-RTTL picks the most uncertain sentences of length 1-10 to fill 10\% of sentences it selects. 

When matching the test distribution (if it is available) S-RTTL provides a way to help deal with dataset shift. Here we apply S-RTTL to match the training data length distribution, though because our test/training split is performed randomly there is no functional difference.

\section{Experiments}

\subsection{Datasets}

We run experiments translating three languages into English, namely Kinyarwanda, Spanish and Swahili. We choose Swahili and Kinyarwanda because they are true low-resource African languages whereas we chose Spanish because it was the original language used to demonstrate RTTL, though we mimic a very low-resource setting in our case. 

Our experiments on Swahili-English uses the SAWA Swahili dataset which consists of data from various domains~\citep{de-pauw-etal-2009-sawa,Pauw2011ExploringTS}. The SAWA dataset is provided on request by the authors \citep{de-pauw-etal-2009-sawa}. In total, there were 272,544 parallel sentences. Pre-processing included: removing missing rows with missing translations and removing sentences with more than 100 words. This clean set consisted of 89,505 parallel sentences. We carried out 5 runs where we split the data into 5 separate folds with each test consisting of 17,901 parallel sentences. Of the remaining sentences, we select 30,000 sentences as our base training set and 1,000 sentences as our validations set. This leaves 40,604 sentences to select from using AL.  With each iteration, we use AL to select 5,000 new sentences and add them to the training set. The validation and test set remain the same for each active learning strategy through all iterations to ensure fair comparison of the methods. Byte Pair Encoding with 4,000 merge operations was applied to the dataset~\citep{sennrich-etal-2016-neural}. 

For the Spanish-English dataset, we utilised Europarl version 7 which  consists of $\sim 2 M$ parallel sentences. For the Kinyarwanda-English dataset, we used the JW300 dataset ~\citep{agic-vulic-2019-jw300} which contains 436,753 parallel sentences. Pre-processing for these 2 datasets included removing sentences with more than 100 words and removing sentences that are the same in both languages. We then randomly chose 148,000 sentences to work with. As with the Swahili-English dataset, we split the data into 5 separate folds. With each test fold, we randomly selected 17,000 sentences as the test set. Thereafter, we concatenate the remaining sentences with the other sentences in the training fold. We randomly select 30,000 sentences as our base training set and 1,000 sentences as our validations set. Contrary to the Swahili-English dataset, due to computational limitations we only conduct one AL iteration where we select 20k sentence out of the 100k remaining pool of training sentences.

\subsection{Training}

We used the transformer-based NMT model JoeyNMT library~\citep{kreutzer-etal-2019-joey}. The configuration for all the models were: 6 layers for both encoder and decoder, 4 heads, embedding dimension of 256, feedforward size of 1,024. We train over 60 epochs, applying Adam optimizer with $\beta_1$ =0.9 and $\beta_2$=0.98 and a learning rate of 0.001.

All training was done on Google Colab which typically provides the Tesla K80 GPU. Every model iteration, both training and testing, was run an average of 4 hours with approximately 12M parameters.

\begin{figure}
    \centering
    \resizebox{\columnwidth}{!}{
        \includegraphics[height=0.27\textheight,width=0.5\textwidth]{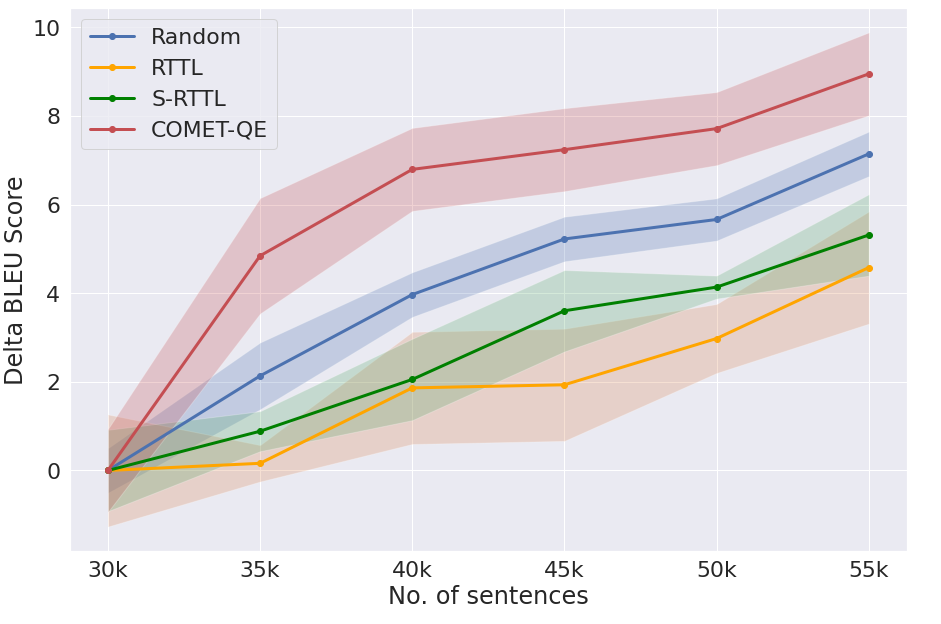}
        }
        \caption{Comparison of the four AL approaches used on Swahili in terms of $\Delta$ BLEU score relative to the baseline of 30,000 sentences. COMET-QE as an AL strategy outperforms the rest of the techniques whereas the RTTL performs the poorest.}
        \label{fig:Result}
\end{figure}

\subsection{Results and Discussion}

\begin{table*}[tb]
\centering
\resizebox{\linewidth}{!}{
\begin{tabular}{l|cccc|cccc}
                                     & \multicolumn{4}{c|}{\textbf{Sentence Length}}                      & \multicolumn{4}{c}{\textbf{Number of symbols}}                               \\ \hline
\textbf{Iteration}                   & \textit{Random} & \textit{RTTL} & \textit{S- RTTL} & \textit{COMET-QE} & \textit{Random} & \textit{ RTTL} & \textit{S-RTTL} & \textit{COMET-QE} \\ \hline
\textit{35k}                         & 11              & 5.3              & 10.2             & 24.7              & 12.4            & 5.5                 & 11.3                     & 29.0              \\
\textit{40k}                         & 10.7            & 9.0              & 10.4             & 16.4              & 12.1            & 9.9                 & 11.5                     & 18.7              \\
\textit{45k}                         & 10.7            & 9.8              & 10.2             & 12.4              & 12.0            & 10.8                & 11.2                     & 14.0              \\
\textit{50k}                         & 10.8            & 11.0             & 10.5             & 9.8               & 12.1            & 12.3                & 11.7                     & 10.9              \\
\textit{55k}                         & 10.9            & 12.7             & 10.8             & 7.7               & 12.3            & 14.5                & 12.1                     & 8.4               \\ \hline
\multicolumn{1}{c|}{\textit{Average}} & 10.82           & 9.56             & 10.42            & 14.2              & 12.18           & 10.6                & 11.56                    & 16.2                 
\end{tabular}
}
\caption{Analysis of the length of sentences and number of symbols in a sentence picked at each iteration of the AL process for Swahili. COMET-QE favors longer sentences while  RTTL favors shorter sentences.  RTTL seems to also pick sentences with a high ratio between sentence length and number of symbols.}
\label{tab:comparison}
\end{table*}

Despite its known limitations we choose the BLEU score ~\citep{papineni-etal-2002-bleu} as a standard metric for comparing the results for the different models. We find that COMET-QE outperforms all of the other methods (RTTL, S-RTTL and random) on all three languages, often by a significant margin. This is shown for Swahili in Figure (\ref{fig:Result}) and for all languages in Table (\ref{tab:Result}). COMET-QE has nearly double the BLEU score gain compared to RTTL for Swahili after adding 25,000 sentences to the baseline. Error contours are estimated 1-$\sigma$ bounds based on our multiple runs. 

Our results show that COMET-QE is a  promising active learning selection strategy in the very low-resource setting. Looking deeper into the sentences that AL with COMET-QE selects (see Table \ref{tab:comparison}), we see that COMET-QE initially selects sentences that are  significantly longer (24.7 words per sentence) with more symbols than any of the other methods. With each iteration it progressively chooses shorter and shorter sentences, ending with just 7.7 words per sentence. 

The exact opposite strategy is chosen by RTTL which starts off selecting short sentences (5.3 words/sentence) and gradually lengthening them, ending with 12.7 words per sentence. Random and S-RTTL choose a middle path, with the average length and symbol complexity being roughly constant with each iteration (about 10.5 words/sentence). 

This same pattern is seen in the number of unique words in each group of 5,000 sentences chosen by the algorithms at each iteration. COMET-QE's initial choice has almost 20,000 unique words, nearly double that of RTTL. In contrast, by the 5th iteration, COMET-QE has halved the number of unique words while RTTL has increased by $\sim 20\%$, see Table (\ref{tab:word}).

\begin{table}
\centering
\resizebox{\columnwidth}{!}{
\begin{tabular}{l|cccc}
Language & Random & RTTL & S-RTTL & COMET-QE \\ 
\hline
Swahili            & $5.7\pm 0.8$  & $3.0\pm 1.2$   & $4.1\pm 0.9$   & $\bf{7.7\pm 1.2}$  \\
Spanish    & $3.2\pm 0.4$       & $2.8 \pm 0.3$    & $1.7 \pm 0.5$   & $\bf{3.5 \pm 0.4}$  \\
Kinyarwanda    & $3.4 \pm 0.3$       & $2.3 \pm 0.3$    & $1.5 \pm 0.5$   & $\bf{5.1 \pm 0.5}$  \\
\hline
\end{tabular}
}
\caption{Gains in BLEU score relative to a 30k sentence baseline for each of Swahili, Spanish and Kinyarwanda with  20k sentences selected using the various AL algorithms. Error estimates are $1-\sigma$ errors over the five runs. The best algorithm is highlighted in bold: in all cases COMET-QE outperforms the other AL techniques. With these small amounts of data RTTL and S-RTTL perform the poorest. The baseline scores for the languages with 30k sentences were: Swahili: $20.8 \pm 0.7$, Spanish: $29.0 \pm 0.2$, Kinyarwanda: $15.7 \pm 1.2$.}
\label{tab:Result}
\end{table}

In Figure (\ref{fig:Result}) we notice that RTTL performs the worst of all the AL strategies, contrary to what was reported in ~\citet{zeng-etal-2019-empirical} and ~\citet{DBLP:journals/corr/abs-2201-05700} where RTTL significantly outperforms random. However,  poor performance of RTTL relative to random selection was also recently observed in \citet{hu-neubig-2021-phrase}. Here we attribute the poor performance of RTTL to the fact that we are exploring the very-low resource setting (only 50k sentences in total). RTTL requires using the learned translation models in both directions, e.g. En-Sw and Sw-En, and hence errors in each direction compound and are amplified. In comparison COMET-QE only requires translation in one direction and therefore one might expect it to perform better in the very low-resource setting. Exploring the selections made by RTTL, we note that it tended to select sentences that are not appropriate because they are dominated by symbols or other noise of no value to learning a translation model. 

Interestingly, S-RTTL improves on RTTL but despite choosing sentences with almost identical average sentence length to the random selection algorithm, still returns a BLEU score about 1.5 lower on average compared to random; see Table (\ref{tab:comparison}).  

\begin{table}
\centering
\resizebox{\columnwidth}{!}{
\begin{tabular}{cllll}
\multicolumn{1}{r}{\textbf{}} & \multicolumn{1}{r}{\textbf{Random}} & \multicolumn{1}{r}{\textbf{ RTTL}} & \multicolumn{1}{r}{\textbf{S-RTTL}} & \multicolumn{1}{r}{\textbf{COMET-QE}} \\ \hline
\textbf{35k}                  & 12,448                               & 10,662                                   & 14,356                                        & 19,995                                 \\
\textbf{40k}                  & 12,253                               & 12,797                                   & 13,375                                        & 15,755                                 \\
\textbf{45k}                  & 12,299                               & 12,709                                   & 12,719                                        & 13,258                                 \\
\textbf{50k}                  & 12,227                               & 12,779                                   & 12,397                                        & 11,589                                 \\
\textbf{55k}                  & 12,394                               & 13,095                                   & 12,107                                        & 9,948  \\  \hline                             
\end{tabular}
}
\caption{Number of unique Swahili words picked at each iteration by each of the algorithms.}
\label{tab:word}
\end{table}

\section*{Conclusions}

We study various active learning strategies in a very low-resource Neural Machine Translation (NMT) setting, specifically using Swahili, Kinyarwanda and Spanish to English as our target languages. Our experiments show that using COMET-QE, a reference-free quality estimation metric, as an Active Learning strategy significantly performs Round Trip Translation Likelihood (RTTL), the current SOTA active learning method as well as random selection. 

COMET-QE initially selects long sentences with a much greater diversity of words than the other algorithms, leading to a BLEU score increase of up to 5 over the 30k baseline for Swahili with just an extra 5,000 sentences added. COMET-QE then progressively chooses shorter sentences with fewer rare words with each successive iteration. Conversely, we find that RTTL, performs worse than random sentence selection. RTTL starts by selecting short sentences and then gradually increases the sentence length in each iteration. To improve its performance, we introduced Stratified RTTL (S-RTTL) which matches the test or training set sentence length distribution. This method leads to  improved performance for Swahili but not for Spanish and Kinyarwanda and was still  worse than random. We attribute the poor performance of RTTL and S-RTTL to the very low-resource setting: the translation models are still weak due to lack of data and since RTTL uses them in both translation directions its estimates of sentence quality are weak. 

This shows that advances of translation quality estimation metrics can be effectively leveraged to improve active learning strategies for very low-resource NMT. 

\section*{Limitations}

The above experiments were conducted on only three languages and limit the study to a very low-resource setting, below 55k sentences. This means that the results may not apply to other languages and investigating the performance of COMET-QE as well as RTTL on other languages and settings is left to future work.

\section*{Acknowledgements}

We wish to thank the Masakhane community for support. We thank Emmanuel Dufourq and the anonymous reviewers for comments and contributions to the draft. This publication was made possible by a grant from Carnegie Corporation of New York (provided through the African Institute for Mathematical Sciences). The statements made and views expressed are solely the responsibility of the authors.

\bibliography{anthology,custom}
\bibliographystyle{acl_natbib}

\end{document}